\documentclass[11pt,a4paper]{article}
\usepackage[hyperref]{acl2020}
\usepackage{times}
\usepackage{latexsym}
\usepackage{array}
\usepackage{multirow}
\usepackage{graphicx}

\usepackage[title]{appendix}
\usepackage{afterpage}
\usepackage{float}

\usepackage{siunitx}
\usepackage{pifont}
\usepackage{xcolor}
\usepackage{amsmath}
\usepackage[T1]{fontenc}
\usepackage{xspace}
\usepackage{multicol}
\usepackage{graphicx}

\newcommand{\OC}{OC\xspace}

\newcommand{\Cat}{\mathcal{C}}
\newcommand{\D}{\mathcal{D}}
\newcommand{\M}{\mathtt{Eff}}
\newcommand{\tick}{\ding{51}}
\newcommand{\fail}{-}

\newcommand{\cProxL}[4]{{#1} \mathrel{{\preceq^{#3}_{#4}}} {#2}}

\newcommand{\ea}[1]{}
\newcommand{\jg}[1]{}
\newcommand{\sm}[1]{}
\newcommand{\jc}[1]{}

\usepackage{soul,color}
\soulregister\cite7
\soulregister\ref7
\soulregister\pageref7
\setstcolor{red}
\setul{}{.2ex}
\usepackage{booktabs}
\newcommand{\CIQ}{\mathrm{CIQ}\xspace}

\newcommand{\prox}{\mathrm{prox}\xspace}
\newcommand{\CEM}{\mathrm{CEM}\xspace}

\newcommand{\Acc}{\mathrm{Acc}\xspace}
\newcommand{\Pre}{\mathrm{Pre}\xspace}
\newcommand{\Rec}{\mathrm{Rec}\xspace}
\newcommand{\UIR}{\mathrm{UIR}\xspace}

\newtheorem{defin}{Definition}

\newtheorem{prop}{Proposition}
\newtheorem{property}{Property}

\usepackage{microtype}

\aclfinalcopy 
\def\aclpaperid{1277} 

\title{An Effectiveness Metric for Ordinal Classification: \\ Formal Properties and Experimental Results}

\author{Enrique Amig\'{o} \\
  UNED \\
  Madrid, Spain \\
  {\small \texttt{enrique@lsi.uned.es}} \\\And
  Julio Gonzalo \\
  UNED \\
  Madrid, Spain \\
  {\small \texttt{julio@lsi.uned.es}} \\\AND
  Stefano Mizzaro \\
  University of Udine \\
  Udine, Italy \\
  {\small \texttt{mizzaro@uniud.it}} \\\And
  Jorge Carrillo-de-Albornoz \\
  UNED \\
  Madrid, Spain \\
  {\small \texttt{jcalbornoz@lsi.uned.es}}%
  }

\begin{document}
\maketitle
\begin{abstract}
In Ordinal Classification tasks, items have to be assigned to classes that have a relative ordering, such as {\em positive}, {\em neutral}, {\em negative} in sentiment analysis. Remarkably, the most popular evaluation metrics for ordinal classification tasks either ignore relevant information (for instance, precision/recall on each of the classes ignores their relative ordering) or assume additional information (for instance, Mean Average Error assumes absolute distances between classes). In this paper we propose a new metric for Ordinal Classification, {\em Closeness Evaluation Measure}, that is rooted on Measurement Theory and Information Theory. 
Our theoretical analysis and experimental results over both synthetic data and data from NLP shared tasks indicate that the proposed metric captures quality aspects from different traditional tasks simultaneously. In addition, it generalizes some popular classification (nominal scale) and error minimization (interval scale) metrics, depending on the measurement scale in which it is instantiated.
\end{abstract}

\section{Introduction}
In Ordinal Classification (OC) tasks, items have to be assigned to classes that have a relative ordering, such as {\em positive}, {\em neutral}, {\em negative} in sentiment analysis. It is different from n-ary classification, because it considers ordinal relationships between classes. It is also different from ranking tasks, which only care about relative ordering between items, because it requires category matching; and it is also different from value prediction, because it does not assume fixed numeric intervals between categories. 

Most research on Ordinal Classification, however, evaluates systems with metrics designed for those other problems. But classification measures ignore the ordering between classes, ranking metrics ignore category matching, and value prediction metrics are used by assuming (usually equal) numeric intervals between categories. 

%

In this paper we propose a metric designed to evaluate Ordinal Classification systems which relies on concepts from Measurement Theory and from Information Theory. The key idea is defining a general notion of closeness between item value assignments (system output prediction vs gold standard class) which is instantiated into ordinal scales but can be also be used with nominal or interval scales. 
Our approach establishes closeness between classes in terms of the distribution of items per class in the gold standard, instead of assuming predefined intervals between classes. 
We provide a formal (Section~\ref{sec:theoEvidence}) and empirical (Section~\ref{sec:exp}) comparison of our metric with previous approaches, and both analytical and empirical evidence indicate that our metric suits the problem best than the current most popular choices. 


\section{State of the Art}
\label{sec:previous}

In this section we first summarize the most popular metrics used in \OC evaluation campaigns, and then discuss previous work on \OC evaluation. 

\subsection{\OC Metrics in NLP shared tasks}

\OC does not match traditional classification, because the ordering between classes makes some errors more severe than others. For instance, misclassifying a {\em positive} opinion as {\em negative} is a more severe error than as a {\em neutral} opinion. Classification metrics, however, have been used for \OC tasks in several shared tasks (see Table~\ref{tab:campaigns}). For instance, Evalita-16 \cite{evalita2016}  uses $F_1$, NTCIR-7 \cite{Kando-08} uses Accuracy, and Semeval-17 Task 4 \cite{SemEval:2017:task4} uses Macro Average Recall. 

\begin{table}[tbp]
\caption{Metrics used for \OC in evaluation campaigns 
\label{tab:campaigns}}
\begin{center}
{\small
\begin{tabular}{@{}l c@{} ccc r c r c r cc@{}}
\toprule
&&\rotatebox{90}{Acc} &\rotatebox{90}{F$_1$}&\rotatebox{90}{AvgRec}&&  \rotatebox{90}{Pearson} && \rotatebox{90}{R/S} &&\rotatebox{90}{MAE$^{M}$} &\rotatebox{90}{MSE}\\
\midrule
NTCIR-7 &&\tick&&&&&&&&& \\
REPLAB-13 &&&&&&&&\tick&&&\\
SEM15-T11&&&&&&&&&&&\tick \\
EVALITA-16&&&\tick&&&&&&&& \\
STS-16  &&&&&&\tick&&&&& \\
SEM17-T4&&&&\tick&&&&&&\tick& \\
\bottomrule
\end{tabular}
}
\end{center}

\end{table}

\OC does not match ranking metrics either: three items categorized by a system as {\em very high/high/low}, respectively, are perfectly ranked with respect to a ground-truth {\em high/low/very\_low}, but yet no single item is correctly classified. However, ranking metrics have been applied in some campaigns, such as R/S for reputation polarity and priority in Replab-2013 \cite{replab2013}.  

\OC has also been evaluated as a value prediction problem -- for instance, SemEval 2015 Task 11 \cite{Ghosh-15} -- with metrics such as Mean Average Error (MAE) or Mean Squared Error (MSE), usually assuming that all classes are equidistant. But, in general, we cannot assume fixed intervals between classes if we are dealing with an \OC task. For instance, in a paper reviewing scale {\em strong\_accept/ accept /weak\_accept / undecided/ weak\_reject/ reject/ strong\_reject}, the differences in appreciation between each ordinal step do not necessarily map into predefined numerical intervals.  

Finally, \OC has been also considered as a linear correlation problem. as in the Semantic Textual Similarity track \cite{Cer-17}. An \OC output, however, can have perfect linear correlation with the ground truth without matching any single value. 

This diversity of approaches -- which do not happen in other types of tasks -- indicates a lack of consensus about what tasks are true Ordinal Classification problems, and what are the general requirements of \OC evaluation. 

\subsection{Studies on Ordinal Classification}

There is a number of previous formal studies on \OC in the literature. First, the problem has been studied from the perspective of loss functions for ordinal regression Machine Learning algorithms. In particular, in a comprehensive work, \citet{Rennie-05} reviewed the existing loss functions for traditional classification and they extended them to \OC. 
Although they did not try to formalize \OC tasks, in further sections we will study the implication of using their loss function for \OC evaluation purposes. 

Other authors analyzed \OC from a classification perspective. For instance, \citet{Waegeman-06}  presented an extended version of the ROC curve for ordinal classification, and \citet{Vanbelle-09}  studied the  properties  of the Weighted Kappa coefficient in \OC. 

Other authors applied a value prediction perspective.  \citet{Gaudette-09}  analysed the effect of using different error minimization metrics for \OC. 
\citet{Baccianella-09}   focused on imbalanced datasets. They imported macro averaging (from classification) to error minimization metrics such as MAE, MSE, and Mean Zero-One Error. 

Remarkably, a common aspect of all these contributions is that {\em they all assume  predefined intervals between categories}. \citeauthor{Rennie-05} assumed, for their loss function, uniform interval distributions across categories. In their probabilistic extension, they assume predefined intervals via parameters in the join distribution model. \citeauthor{Waegeman-06}  explicitly assumed that ``the misclassification costs are always proportional to the absolute difference between the real and the predicted label''. The predefined intervals are defined by \citeauthor{Vanbelle-09} via weighting parameters in Kappa. The MAE and MSE metrics compared by \citeauthor{Gaudette-09}  also assume predefined (uniform) intervals. Finally, the solution proposed by \citeauthor{Baccianella-09}  is based on ``a sum of the classification errors across classes''.

In our opinion, assuming and adding intervals between categories to estimate misclassification errors violates the notion of ordinal scale in Measurement Theory \cite{StevensMeasurement}, which establishes that intervals are not meaningful relationships for ordinal scales. Our measure and our theoretical analysis are meant to address this problem.

\section{Closeness Evaluation Measure (CEM)}
\label{sec:measures}


\begin{figure*}[tb]
\centering
\begin{tabular}{@{}c@{}c@{}}
\includegraphics[width=.49\linewidth]{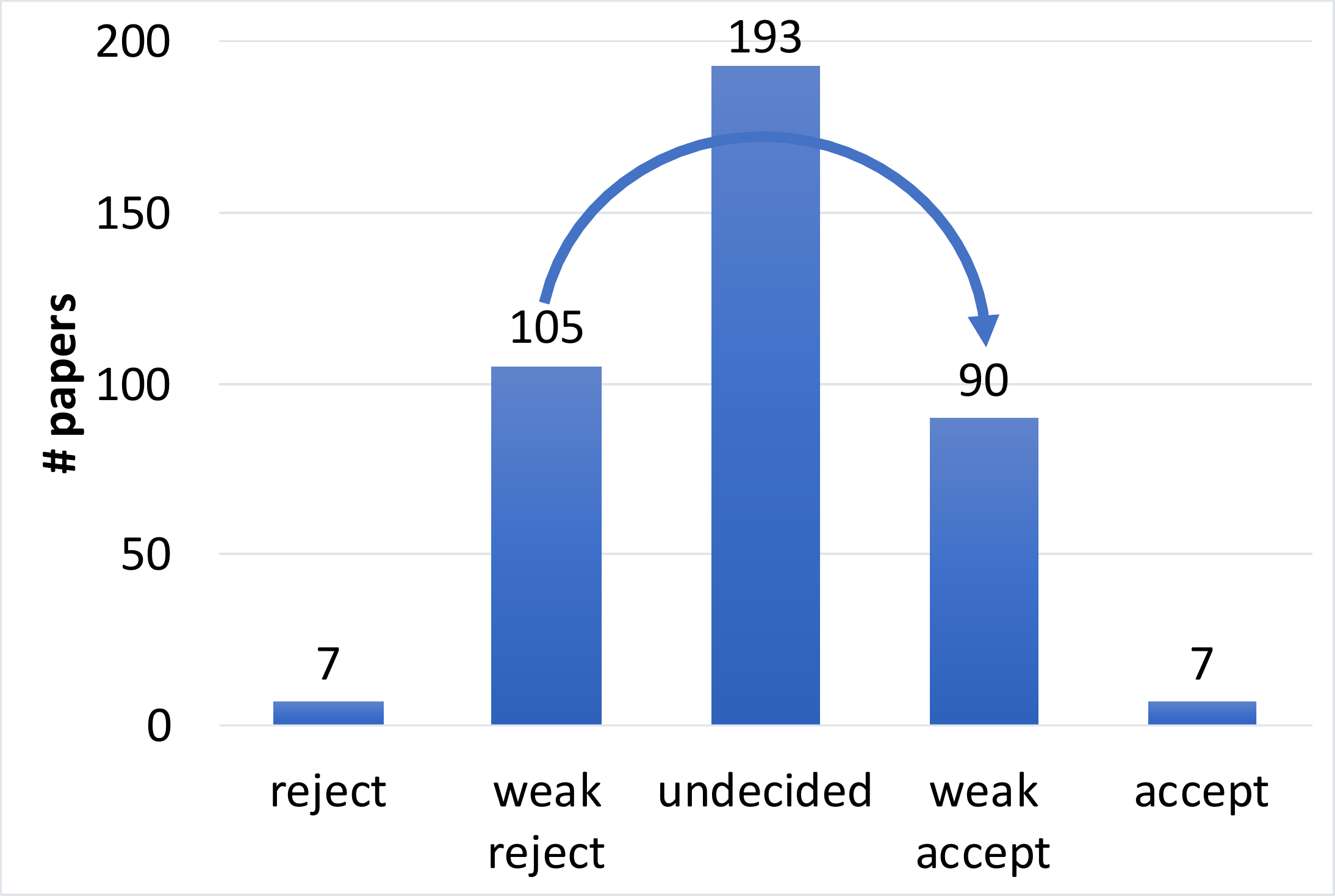} &
\includegraphics[width=.49\linewidth]{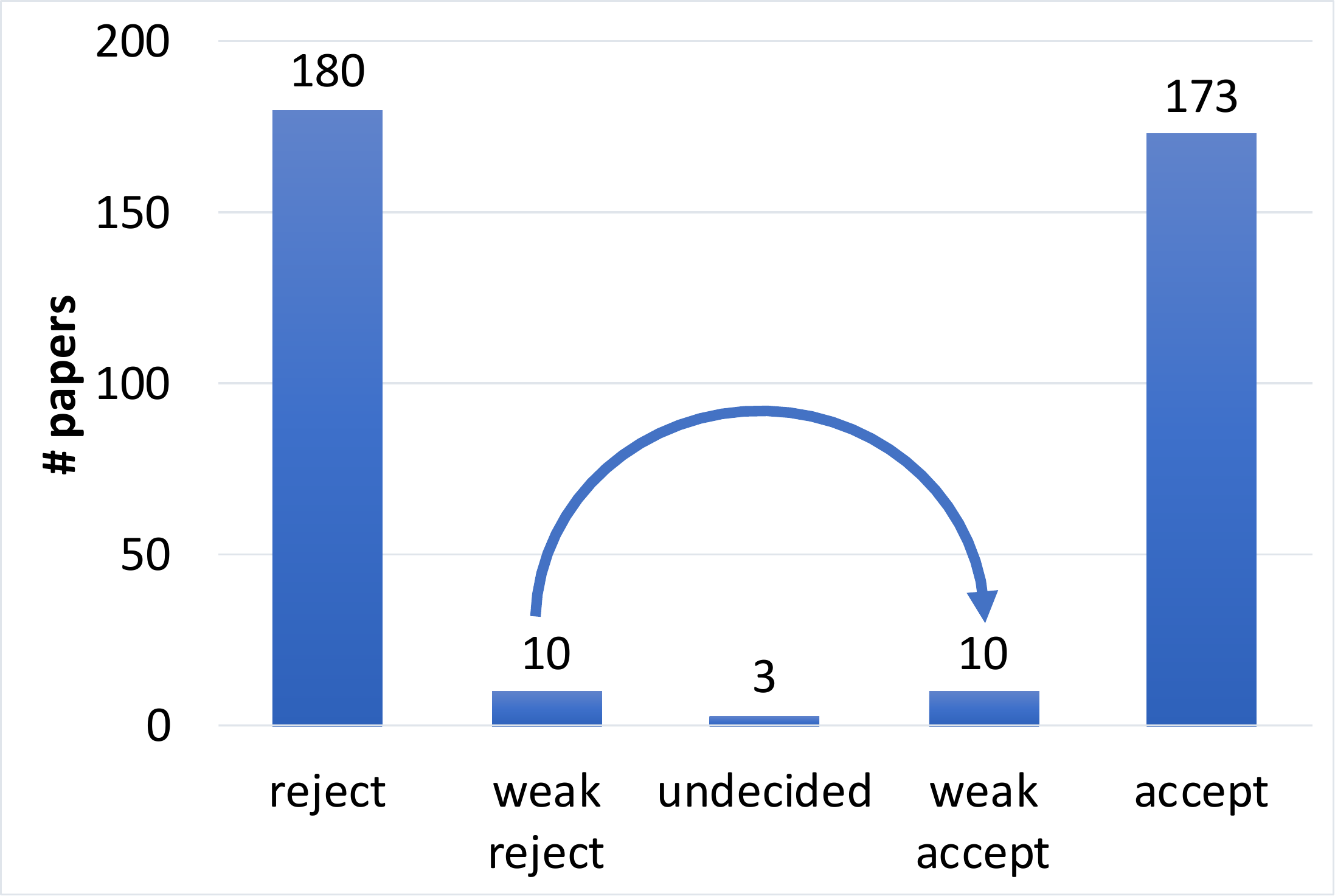}
\end{tabular}
\caption{In the left distribution, {\em weak accept} vs. {\em weak reject} would be a strong disagreement between reviewers (i.e., the classes are distant), because in practice these are almost the extreme cases of the scale (reviewers rarely go for {\em accept} or {\em reject}). In the right distribution the situation is the opposite: reviewers tend to take a clear stance, which makes {\em weak accept} and {\em weak reject} closer assessments than in the left case.}
     \label{fig:distributions}
\end{figure*}

\subsection{Measure Definition} 

Evaluation metrics establish proximity between a system output and the gold standard \cite{monster20}. In ordinal classification we have to compare the classes assigned by the system with the true classes in the gold standard. 




A key idea in our metric is to establish a notion of {\em informational closeness} that depends on how items are distributed in the rank of classes. The idea is that two items $a$ and $b$ are {\em informationally close} if the probability of finding an item between the two is low. As an example, Figure~\ref{fig:distributions} illustrates the intuition of how item distribution affects informational closeness in the context of paper reviewing. This is similar in spirit to, for instance, comparing the quality of two journals according to their quartiles in the rank of journals of comparable topics. With this notion of informational closeness, proximity between classes adapts to the way in which classes are used in a given dataset. 

This idea of {\em informational closeness} can be implemented using Information Theory: the more unexpected it is to find an item between $a$ and $b$, the more information such event provides, and the more $a$ and $b$ are informationally closer. 
Let $P(\cProxL{x}{a}{b}{\mathtt{ORD}})$ be the probability that, sampling an item $x$ from the space of items, $x$ is closer to $b$ than $a$ in the ordinal scale of classes. Then we can define {\em Closeness Information Quantity} (CIQ) between $a$ and $b$ as the {\em Information Quantity} of the event $\cProxL{x}{a}{b}{\mathtt{ORD}}$, as follows:
\begin{equation}
    \CIQ^\mathtt{ORD}(a,b) \equiv - \log (P(\cProxL{x}{a}{b}{\mathtt{ORD}})).
\end{equation}


Let us now apply this concept for the evaluation of system outputs. Let $\D$ be the item collection, $\Cat=\{c_1, \ldots, c_n\}$ a set of sorted classes such that $c_1<c_2< \ldots <c_n$, and $g,s:\D\longrightarrow \Cat$ the gold standard and a system output. Given the classes $g(d),s(d)$ assigned to an item $d\in \D$ by the gold standard and the system output, $\CIQ^\mathtt{ORD}(s(d),g(d))$ measures the closeness between the assigned class and the gold standard class:
$$\CIQ^\mathtt{ORD}(s(d),g(d)) = - \log (P(\cProxL{x}{s(d)}{g(d)}{\mathtt{ORD}})).$$

Our proposed evaluation measure consists in adding CIQ values for all items $d \in \D$, and normalizing the sum by its maximal value, which is the one obtained by a system output that matches the gold standard perfectly. This is what we call {\em Closeness Evaluation Measure}, $\CEM^\mathtt{ORD}$:
%
$$\CEM^\mathtt{ORD}(s,g)=\frac{\sum_{d\in\D}\CIQ^\mathtt{ORD}(s(d),g(d))}{\sum_{d\in\D}\CIQ^\mathtt{ORD}(g(d),g(d))}.$$


In an ordinal scale, the condition $\cProxL{x}{a}{b}{\mathtt{ORD}}$ ($x$ is closer to $b$ than $a$) implies that $x$ is between $a$ and $b$ ($a \geq x \geq b$ or $a \leq x \leq b$). 
Therefore, if $n_i$ is the amount of items assigned to class $c_i$ in the gold standard, and $N$ is the total amount of items, the formula above turns into:
$$\CEM^\mathtt{ORD}(s,g)=\frac{\sum_{d\in\D}
\prox(s(d),g(d))}{\sum_{d\in\D} \prox(g(d),g(d))}$$
%
where $\prox(c_i,c_j)=-\log\left(\frac{\frac{n_i}{2}+\sum_{k=i+1}^j n_k}{N}\right)$.

Note that the term $\prox(c_i,c_j)$, which is the core of the metric, reflects the {\em informational closeness} that the metric assigns to a pair of classes $c_i,c_j$. Note also that half of the ties (elements in the class $i$) are included in the computation. Every time the system assigns the class $c_i$ and the ground truth is $c_j$, the contribution of that assignment to the final value of $\CEM^\mathtt{ORD}$ is proportional to the informational closeness between both classes.



As an example, let us consider the two ground truth distributions in Figure \ref{fig:distributions}. The proximity between the classes {\em weak\_accept} and {\em weak\_reject} for the left distribution is:

{\footnotesize
$$-\log\left(\frac{90/2+193+105}{402}\right)=0.23$$
}
and for the right distribution is:

{\footnotesize
$$-\log\left(\frac{10/2+3+10}{376}\right)=4.38.$$
}


A mistake between these two classes is more heavily penalized by the metric in the left distribution. Note also that correct predictions have different weights -- $\prox(c_i,c_i)$ -- which are higher for infrequent classes. For instance, a correct guess for a {\em reject} ground truth in the left distribution has a weight of $\prox$({\em reject,reject})$=6.84$, because it is a rare class (7/402 items); but a correct guess for an {\em undecided} item has only a weight of $2.06$ because the class is very frequent in the ground truth (193/402 items). This is an effect of using Information Theory to characterize closeness: an infrequent class has more information than a frequent class.

Overall, $\CEM^\mathtt{ORD}$ rewards exact matches, considers ordinal relationships, and does not assume predefined intervals between classes (instead, intervals depend on the distribution of items into classes in the gold standard). Appendix A shows detailed examples of how to compute $\CEM^\mathtt{ORD}$ from the confusion matrix for a system output.

\subsection{Formalization of $\CEM$ on Different Scales}
\label{sec:theo}

We have specified our measure $\CEM^\mathtt{ORD}$ at ordinal scale to address \OC tasks, but it could be used at any scale. In this section we briefly investigate this generalization. 
In Measurement Theory, at least in \citeauthor{StevensMeasurement}'s model (\citeyear{StevensMeasurement}), all measures map items to real numbers, and measurement equivalence at different scales is determined by \textit{permissible transformation functions}. Permissible transformations are bijective functions in nominal scale $(\mathcal{F}_{\mathtt{NOM}})$, strictly increasing functions in ordinal scale $(\mathcal{F}_{\mathtt{ORD}})$, and linear functions for the interval scale $(\mathcal{F}_{\mathtt{INT}})$. 

Starting from the notion of $|a-b|$ as the standard algebraic distance between numbers, we define \textit{closeness} at a certain measurement scale $\mathtt{T}$ if it fits for at least one permissible transformation in $\mathcal{F}_{\mathtt{T}}$.

\begin{defin}[Closeness for a Scale Type]\label{def:lcloseness} Being three numbers x, a, and b, we say that   $x$  is   closer to  $b$ than $a$, $(\cProxL{x}{a}{b}{\mathtt{T}})$
for a certain scale type $\mathtt{T}$,
if  and only if:
\begin{equation*}
\exists f \in \mathcal{F}_{\mathtt{T}}\left(
|f(x)-f(b)|\le|f(a)-f(b)|
\right).
\end{equation*}
\end{defin}

The conditions for $\cProxL{x}{a}{b}{\mathtt{T}}$ at ordinal scale $(\mathtt{T}=\mathtt{ORD})$ are $(b\ge x \ge a)   \vee (a \ge x \ge b)$ (see proof in the supplementary material). That is, at ordinal scale, $x$ must be located between $a$ and $b$ to be closer to $a$ than $b$. The condition for nominal  scale $(\mathtt{T}=\mathtt{NOM})$ is  $ \left(b=x \vee b\neq a\right)$. At interval scale $(\mathtt{T}=\mathtt{INT})$, the condition matches the standard algebraic closeness between numbers:  $(|b-x|\le|b-a|)$.

We can generalize $\CIQ^\mathtt{ORD}$ and $\CEM^\mathtt{ORD}$ to consider closeness at any scale $\mathtt{T}$, simply replacing $\cProxL{x}{a}{b}{\mathtt{ORD}}$ with $\cProxL{x}{a}{b}{\mathtt{T}}$. We denote these generalizations as $\CIQ^\mathtt{T}$, $\CEM^\mathtt{T}$. The $\CEM^\mathtt{T}$ metric generalizes some of the most popular metrics in classification.


\begin{prop}
\label{prop:AccLink}
Assuming that categories in $g$ follow a uniform distribution,  then Accuracy is proportional to CEM at nominal scale. Formally, whenever $P(g(d)=c)$ is equal for all categories $c\in\Cat$, then:
$$\Acc(s,g)\propto \CEM^\mathtt{NOM}(s,g).$$
\end{prop}
Macro Average Accuracy can be also defined by aggregating $\CIQ^\mathtt{NOM}(s(d),g(d))$ in the corresponding manner. Also, under the same statistical assumptions, Precision and Recall for a category $c$ can be defined in terms of aggregated $\CIQ$s of items in the  system or gold category respectively.
\begin{prop}
\label{prop:PreRecLink} Whenever $P(g(d)=c)$ is equal for all categories $c\in\Cat$,  then:
\begin{align*}
\Pre_{g,c}(s)\propto & \sum_{d\in\D:s(d)=c} \CIQ^\mathtt{NOM}(s(d),g(d))\\
\Rec_{g,c}(s)\propto & \sum_{d\in\D:g(d)=c} \CIQ^\mathtt{NOM}(s(d),g(d)).
\end{align*}
Exact match between Precision, Recall and the $\CIQ$ aggregation is achieved when values are normalized with respect to the maximum.
\end{prop}

On the other hand, if we do not assume a uniform distribution of items into classes in the gold standard, then we obtain a classification metric CEM$^\mathtt{NOM}(s,g)$ which gives more (logarithmic) weight to errors in infrequent classes. 

Finally, at interval scale, CEM$^\mathtt{INT}$ would be equivalent to a logarithmic version of MAE whenever items are uniformly distributed across classes. 

We leave a more detailed formal and empirical analysis of CEM at other scales for future work, as it is not the primary scope of this paper. 
\section{Theoretical Evidence}
\label{sec:theoEvidence}

\begin{figure}[tb]
      \centering
        \includegraphics[width=0.45\textwidth]{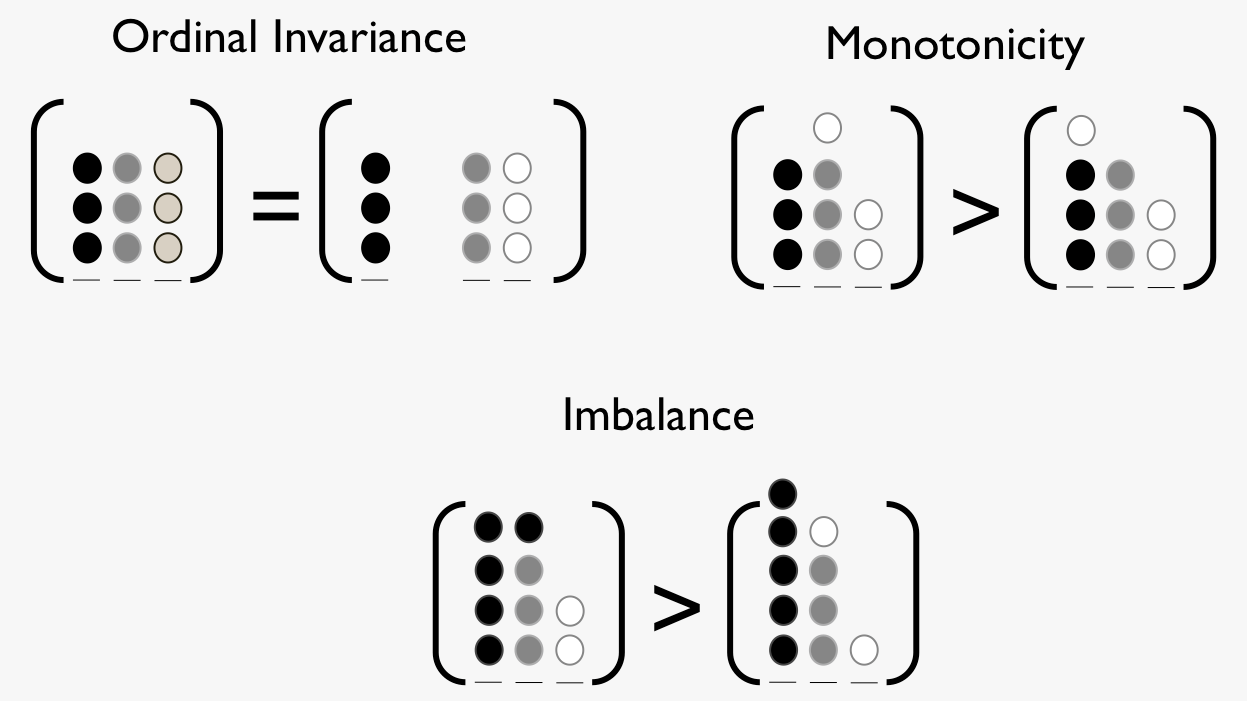}
      \caption{ Illustration of desirable formal properties for Ordinal Classification. Each bin is a system output, where columns represent ordered classes assigned by the system, and colors represent the items' true classes, ordered from black to white. "=" means that both outputs should have the same quality, and ">" that the left output should receive a higher metric value than the right output\label{figProp}.}
\end{figure}

Following a methodology previously applied for Classification \cite{Sebastiani15,Solokova-06}, Clustering \cite{Dom-01,Meila-03,Amigo-09}, and document ranking tasks \cite{moffat:2013,amigo2013general}, here we  define a formal framework for \OC via desirable properties to be satisfied, which are illustrated in Figure~\ref{figProp} and introduced below.

\subsection{Metric Properties}



The first property states that an effectiveness metric $\M(s,g)$ should not assume predefined intervals between classes, i.e., it should be invariant under permissible transformation functions at ordinal scale.
\begin{property}
[Ordinal Invariance] An effectiveness metric $\M$ has ordinal invariance if it is invariant under strictly increasing functions $f_{\mathtt{ORD}} \in \mathcal{F}_{\mathtt{ORD}}$ applied to both the system output and the gold standard:
$$\M(s,g)=\M(f_{\mathtt{ORD}}(s),f_{\mathtt{ORD}}(g)).$$
\end{property}
For instance,  $\M((1,2,2),(1,2,3))$ should be equivalent to $\M((11,24,24),(11,24,39))$, by considering the (strictly increasing) permissible transformation function  $f_{\mathtt{ORD}}(x)=10x+x^2$. 

\begin{table}[tbp]
\caption{Constraint-based Metric Analysis \label{cons} 
}
\begin{center}
\small
\begin{tabular}{@{}l@{ }l@{}c@{ }c@{ }c@{}}
\toprule
&  & \multicolumn{3}{c}{Constraints}\\
\multicolumn{1}{@{}l@{}}{Metric family} & \multicolumn{1}{@{}l}{Metrics} &Ord.   &Ord. &Imb.\\
 &  & Inv.  &Mon. &\\
\midrule
&Acc&\tick&\fail&\fail\\
Classification  &Acc with n&\tick&\fail&\fail\\
Metrics&Macro Avg Acc, Cohen's $\kappa$&\tick&\fail&\tick\\
&F-measure avg. across classes&\tick&\fail&\tick\\
\addlinespace
&MAE, MSE&\fail&\tick&\fail\\
Value &Macro Avg. MAE/MSE&\fail&\tick&\tick\\
Prediction&Weighted $\kappa$&\fail&\tick&\tick\\
&Rennie \& Srebro loss function&\fail&\tick&\fail\\
 &Cosine  similarity &\fail&\tick&\fail\\
\addlinespace
&Linear correlation&\fail&\fail&\fail\\
Correlation  &Ordinal: Kendall (tau-b), Spea. &\tick&\fail&\tick\\
 Coefficients & Kendall-(Tau-a) &\tick&\fail&\fail\\
&Reliability and Sensitivity&\tick&\fail&\tick\\
\addlinespace
Clustering &MI, Purity and Inv. Purity&\tick&\fail&\tick\\
\addlinespace
Path based& Ordinal Classification Index&\tick&\fail&\fail\\
\addlinespace
 &  CEM$^{\mathtt{NOM}}$&\tick&\fail&\tick\\
  CEM& CEM$^{\mathtt{INT}}$&\fail&\tick&\tick\\
  & {\bf CEM}$^{\mathtt{ORD}}$&\tick&\tick&\tick\\
 \bottomrule
\end{tabular}
\end{center}
\end{table}

Although we can not compare intervals at ordinal scale, we know, e.g., that ``neutral'' is closer to ``positive'' than ``negative''. Therefore we need another property to verify  monotonicity  with respect to  category closeness.

\begin{property}[Ordinal Monotonicity] Changing system predictions closer to the true category should result in a metric increase:
%
%
\begin{equation*}
    \begin{split}
        &\mbox{If}\ \ \exists d.(s(d)\neq s'(d)) \wedge \\ & (\forall d . ((s(d)>s'(d)\ge g(d)) \vee  (s(d)=s'(d))))\\
        &\mbox{then}\ \ \M(s',g)>\M(s,g).
    \end{split}
\end{equation*}
\end{property}

The formalization of ordinal monotonicity states that if all predictions by system $s'$ are better or equal than predictions by $s$, and at least one is strictly better, then the metric score of $s'$ must be higher. 

Finally, in order to manage the effect of imbalanced data sets, another desirable property is that an item classification error in a frequent class should have less effect than a classification error in a small class \cite{Fatourechi-08}.  In order to formalize this property, we  use $g_{d\rightarrow c}$ to denote the result of moving the item $d$ to the class $c$ in the gold standard.

\begin{property}[Imbalance] Distancing items from a small class has more effect than distancing items from a large class. Let $(c_1,c_2,c_3)$ be three contiguous classes such that $c_1$ is larger than $c_3$, and $d_1,d_3$ two items such that $g(d_1)=c_1$ and $g(d_3)=c_3$. Then
\begin{equation*}
    \M(g_{d_1\rightarrow c_2},g)>\M(g_{d_3\rightarrow c_2},g).
\end{equation*}
\end{property}

\subsection{Metric Analysis}
Table~\ref{cons} displays the properties satisfied by metrics grouped by families.\footnote{See the supplementary material for proofs and counter examples where appropriate.}
{\bf Classification metrics} are ordinal invariant, but they do not satisfy ordinal monotonicity.  Attempts to mitigate this limitation include (i) Accuracy at n \cite{Gaudette-09} which relaxes Accuracy with an ordinal margin error, and (ii) ignoring the neutral class \cite{semeval14}. However, both approaches are insensitive to some types of error.  Some classification metrics such as MAAC, Cohen's Kappa or F-measure averaged across classes satisfy the imbalance constraint.

The most popular {\bf Value Prediction} metrics  are Mean Absolute Error (MAE) and Mean Square Error (MSE). They both assume a predefined fixed numerical value for each category. Therefore, ordinal invariance is violated. 
The imbalance property is satisfied by the Macro Average versions MAE$^m$ MSE$^m$ \cite{Baccianella-09}. The weighted Kappa can be monotonic whenever the accumulated weights are consistent with the ordinal structure \cite{Vanbelle-09}. In addition, it can satisfy imbalance depending on the weighting scheme. However, ordinal invariance is not satisfied. 
The loss function for ordinal classification proposed by \citet{Rennie-05} is, in the same way as MAE, grounded on category differences, and therefore does not satisfy ordinal invariance. Finally, the cosine similarity has also been employed to evaluate \OC \cite{Ghosh-15}, where documents are dimensions and categories are vector values. Just like any other geometric measure, it is not ordinal invariant and it does not satisfy imbalance. 

In general, {\bf correlation coefficients} do not satisfy  monotonicity,  given that exact matching of gold standard values is not required to achieve the maximum score. Unlike linear correlation, ordinal correlation coefficients (i.e., Kendall or Spearman) are  ordinal invariant.  Kendall can be computed in different ways depending on how ties are managed. In Tau-a, only discordant pairs are considered ($g(d_1)>g(d_2)$ and $s(d_1)<s(d_2)$) and imbalance is not satisfied. The most popular Kendall coefficient approach (Tau-b) and Spearman both satisfy imbalance. Pearson coefficient does not, due to the interval effect. 
Reliability and Sensitivity metrics, which extend the clustering metric BCubed, are essentially an ordinal correlation metric, being invariant but failing in monotonicity, with the advantage of satisfying imbalance due to the precision/recall notions.  

 By definition, {\bf clustering  metrics} are ordinal invariant, because they are not affected by the cluster of category descriptors. In addition, most of them,  such as Mutual Information (MI) or Purity and Inverse Purity, satisfy imbalance. However, they are not ordinal monotonic, given that they do not consider any ordinal relationship between categories.

Finally, we must include the approach by \citet{Cardoso-11}, a path based metric called Ordinal Classification Index which is designed specifically for \OC problems. This is a metric that integrates aspects from the previous three metric families, including 
two parameters $\beta_1$ and $\beta_2$ to combine different components. Therefore, this metric can capture the different quality aspects involved in the \OC process. However, the metric inherits the lack of invariance of MAE and MSE when computing the ordinal distance between categories, and monotonicity can be violated depending on the effect of discordant item pairs.

The table ends with our proposed metric $\CEM$, which is either a classification, error minimization, or \OC metric depending if it is instantiated into nominal ($\CEM^{\mathtt{NOM}}$), interval ($\CEM^{\mathtt{INT}}$), or ordinal measurement scale ($\CEM^{\mathtt{ORD}}$). $\CEM^{\mathtt{ORD}}$ is the only metric that satisfies the three properties, provided that there are no empty classes in the gold standard (see Appendix A.2).

\section{Empirical Study}
\label{sec:exp}

Meta-evaluating metrics is not straightforward. A common criterion is robustness, defined as consistence (correlation) of system rankings across data sets. However, although robustness is relevant -- and we do report it at the end of this section -- it does not reflect to what extent a metric captures the quality aspects of systems.

As many authors have pointed out, an \OC metric should capture diverse aspects of systems: class matching, ordering, and imbalance. In our experiments, in addition to robustness, we select three complementary metrics, each focused on one of these partial aspects, and we evaluate to what extent existing \OC metrics are able to capture all these aspects simultaneously.

The selected metrics are: (i) Accuracy, as a partial metric which captures class matching; (ii)~Kendall's correlation coefficient Tau-a (without counting ties), in order to capture class ordering\footnote{
en.wikipedia.org/wiki/Kendall\_rank\_correlation\_coefficient}; and (iii) Mutual Information (MI), a clustering metric which reflects how much knowing the system output reduces uncertainty about the gold standard values. This metric  accentuates the effect of small classes (imbalance property). 

\subsection{Meta-evaluation Metric}

In order to quantify the ability of metrics to capture the aspects reflected by these three metrics, we use the Unanimous Improvement Ratio (UIR) \cite{UIR}. While \textit{robustness} focuses on consistence across data sets, UIR focuses on consistence across metrics.  It essentially counts in how many test cases an improvement is observed for all metrics simultaneously. Being $\mathcal{M}$ a set of metrics, and $\mathcal{T}$ a set of test cases, and $s_t$ a system output for the test case $t$, the Unanimous Improvement Ratio $\UIR_\mathcal{M}(s,s')$  between two systems is defined as:
$$\frac{\big|\big\{t\in \mathcal{T}:s_t\ge_\mathcal{M} s'_t\big\}\big|- \big|\big\{t\in \mathcal{T}: s'_t\ge_\mathcal{M} s_t\big\}\big|}{\big|\mathcal{T}\big|},$$
where $s_t\ge_\mathcal{M} s'_t$ represents that system $s$ improves system $s'$, on the test case $t$, unanimously for every metric:
$$s_t\ge_\mathcal{M} s'_t\equiv\big(\forall m \in \mathcal{M}\big(m(s_t)\ge m(s'_t\big)\big).$$

Therefore, UIR reflects to what extent a system outperforms another system for several metrics simultaneously. Then, we define our meta-evaluation measure \textit{Coverage} for a single metric $m$ as the Spearman correlation (over system output pairs $s,s'$ in the set of system outputs) between differences in $m$ and unanimous improvements over the reference metric set. Being $\mathcal{M}$ the reference metric set:\footnote{We use the non parametric coefficient Spearman instead of Pearson.  This focuses the meta-evaluation on system score ordering rather than particular scale properties of metrics.}
$$\mbox{Cov}_{\mathcal{M}}(m)=\mbox{Spea}\bigg(m(s)-m(s'), \UIR_\mathcal{M}(s,s')\bigg).$$
The more the coverage of a metric $m$ is high with respect to a reference metric set $\mathcal{M}$, the more an improvement according to $m$ reflects all quality aspects represented by $\mathcal{M}$.  

\subsection{Compared Metrics}

We evaluate the  coverage of $\CEM^\mathtt{ORD}$ and other metrics with respect to the reference metric set Accuracy, Kendall, and MI. In the empirical study we have considered most metrics used in practice to evaluate \OC problems; we have excluded a few metrics which are included in the theoretical study, either because they have not been used previously to evaluate \OC problems (such as clustering metrics) or because they have internal parameters and therefore a range of variability that requires a dedicated study (such as weighted Kappa and Ordinal Index).  
In order to check the need for the logarithmic scaling in $\CEM^\mathtt{ORD}$ (which comes from the application of \emph{Information Quantity}), we also include an alternative metric $\CEM^\mathtt{ORD}_{flat}$, which is similar to CEM but without the logarithmic scaling. 
\begin{table*}[tb]
  \centering
\caption{Metric Coverage: Spearman Correlation between single metrics and the UIR combination of Mutual Information, Accuracy, and Kendall across system pairs in both the synthetic and real data sets. \label{tab:exp1} }
\small
\begin{tabular}{l@{ }l  c@{ }c@{ }c@{ }c@{ }c@{ }c   @{ }c   c@{ }  c@{ }c c@{ }c@{ }c}
\toprule
&&\multicolumn{6}{c}{Synthetic data} & \multicolumn{7}{c}{Real data}\\
\cmidrule{3-8}\cmidrule{10-15}
&  & all & minus  & minus & minus & minus &  minus && Replab & \multicolumn{2}{c}{SEM-2014}  & \multicolumn{3}{c}{SEM-2015}\\
&  & systems & $s_{Rand}$ & $s_{prox}$ & $s_{maj}$ & $s_{tDisp}$ & $s_{oDisp}$ && 2013 & T9-A & T9-B & T10-A & T10-B & T10-C\\
\cmidrule{3-8}\cmidrule{10-15}
Reference & Accuracy & 0.81 & 0.77 & 0.78 & 0.78 & 0.94 & 0.77&& 0.75 & 0.90 & 0.98 & 0.85 & 0.94 & 0.80\\
metrics in & Kendall & 0.84 & 0.81 & 0.82 & 0.82 & 0.93 & 0.82 && 0.88 & 0.94 & 0.98 & 0.84 & 0.97 & 0.88\\
 UIR & MI & 0.84 & 0.82 & 0.84 & 0.82 & 0.93 & 0.82 && 0.91 &{\bf 0.97 }& 0.99 & 0.93 & 0.98 & 0.93\\
\addlinespace
 & F-measure & 0.83 & 0.80 & 0.82 & 0.81 & 0.93 & 0.81 && 0.66 & 0.90 & 0.98 & 0.91 & 0.98 & 0.92\\
Classification & MAAC & 0.83 & 0.81 & 0.82 & 0.79 & 0.91 & 0.81 && 0.84 & 0.86 & 0.97 & 0.84 & 0.95 & 0.82\\
metrics & Kappa & 0.81 & 0.78 & 0.79 & 0.77 & 0.94 & 0.77 && 0.44 & 0.95 & {\bf 0.99} & 0.93 & 0.98 & {\bf 0.97}\\
 & Acc with 1 & 0.79 & 0.75 & 0.77 & 0.80 & 0.85 & 0.79 && 0.23 & 0.82 & 0.60 & 0.31 & 0.35 & -0.19\\
\addlinespace
 & MAE & 0.84 & 0.82 & 0.83 & 0.87 & 0.86 & 0.84 && 0.81 & 0.96 & 0.95 & 0.95 & 0.87 & 0.56\\
Error & MAE$_m$ & 0.74 & 0.73 & 0.74 & 0.80 & 0.76 & 0.73 && 0.73 & 0.95 & 0.88 & 0.91 & 0.74 & 0.30\\
minimization & MSE & 0.89 & 0.87 & 0.87 & 0.88 & 0.93 & 0.88 && 0.28 & 0.87 & 0.98 & 0.63 & 0.97 & 0.93\\
 & MSE$_m$ & 0.83 & 0.80 & 0.80 & 0.82 & 0.90 & 0.83 & & 0.10 & 0.85 & 0.94 & 0.48 & 0.91 & 0.52\\
\addlinespace
Correlation & Pearson & 0.77 & 0.79 & 0.74 & 0.73 & 0.83 & 0.79 && 0.91 & {\bf 0.97} & 0.98 & 0.96 & 0.97 & 0.79\\
coefficients & Spearman & 0.72 & 0.67 & 0.69 & 0.77 & 0.76 & 0.70 && 0.07 & 0.96 & 0.98 & 0.97 & 0.98 & 0.80\\
\addlinespace
Measurement & CEM$^{\mathtt{ORD}}$ & {\bf 0.91 }& {\bf  0.89 }& {\bf 0.90 }& {\bf 0.90 }& {\bf 0.95 }& {\bf 0.89} && {\bf 0.94} & 0.96 & {\bf 0.99} & {\bf 0.98} &{\bf 0.99} & 0.96\\
theory & CEM$^{\mathtt{ORD}}_{flat}$ & 0.87 & 0.84 & 0.86 & 0.88 & 0.89 & 0.87 & & 0.82 & 0.96 & 0.96 & 0.94 & 0.92 & 0.65\\
\bottomrule
\end{tabular}
\end{table*}

\subsection{Experiments on Synthetic Data}
In order to play with a representative and controlled amount of classes and distributions, we first experiment with synthetic data. Let us consider a synthetic dataset with 100 test cases and 200 documents per test case, classified into 11 categories. In order to study different degrees of imbalance, we assign ground truth labels to documents according to a normal distribution with average 4 and a typical deviation between 1 and 3. The imbalance grade (deviation) varies uniformly across topics. The majority class is therefore the fourth class.\footnote{We selected class 4 instead of 6  in order to have asymmetry in the distribution.} Finally, we discretize the resulting values into their closest category in $\{1,2,\ldots ,11\}$.

We generate synthetic system outputs according to the  following behaviour: each system makes mistakes in a certain ratio $r$ of value assignments, where $r\in\{0.1,0.2,\ldots ,0.9,1\}$. Then we distinguish between five kinds of mistakes, thus obtaining $10\times 5$ possible system configurations. The five alternative mistakes are:
\begin{enumerate}    
    \item {\bf Majority  class assignment:} Assign the most frequent category: $s_{maj}(d)=4$.
    \item {\bf Random assignment:} Assign classes randomly: $s_{rand}(d)=v$ with $v\sim U(1,11)$.
    \item {\bf Tag displacement:} Assign the next category: $s_{tDisp}(d)=g(d)+1$.    
    \item {\bf Ordinal displacement:} Being $ord(d)$ the ordinal position of $d$ in  a  sorting of documents  in concordance with category values $(g(d)>g(d')\Rightarrow ord(d)>ord(d'))$, the system displaces the document $\frac{n}{10}$ positions:
    {\small
    $$s_{oDisp}(d_i)=g\left(d':ord(d')=ord(d)+\frac{n}{10}\right).$$
    }
    \item {\bf Proximity assignment:} The assignment is closer to the gold standard than a random one: it assigns a category between a randomly selected one and the gold standard:
    {\small
    $$s_{prox}(d)=g\left(d':ord(d')=\frac{ord(d)+rPos}{2}\right)$$
    }
    with $rPos\sim U(1,n)$ (a random position between $1$ and $n$).  
\end{enumerate}

We discretize  the resulting values in the same way than the gold standard.  The synthetic outputs are designed to produce trade-offs between evaluation metrics. For instance, a total displacement $(s^{r=1}_{tDisp})$ achieves the maximal Kendall correlation but the lowest Accuracy. On the contrary, a 30\% of random assignments $s_{\{r=0.3,rand\}}$ can decrease substantially the ordinal relationships, but keeping a 70\% of Accuracy. Also,  $s^{r=0.3}_{rand}$  outperforms
$s^{r=0.5}_{prox}$ in terms of accuracy, but not necessarily in terms of error minimization metrics. Finally, $s^{r=0.3}_{rand}$ can be outperformed by $s^{r=0.4}_{maj}$ given that the second system assigns documents to the majority class, but not in terms of MI, which accounts for the imbalance effect.

Table~\ref{tab:exp1} (left part) shows the results. The metric coverage can vary substantially when changing the distribution of systems. For this reason, we first consider every synthetic output and then we repeat the experiment removing each of the system types. 
As the table shows, $\CEM^\mathtt{ORD}$ improves all other metrics, including the individual metrics used as a reference via UIR (MI, Kendall, and Accuracy).  Note that the  flat (not logarithmic) version $\CEM^\mathtt{ORD}_{flat}$ performs systematically worse than the original metric, which supports the use of the logarithmic, information-theoretic formula to compute similarity. 

\subsection{Experiments on NLP shared tasks}

Let us now study how metrics behave with actual data from evaluation campaigns, where we cannot control the amount and types of error. We use data from six \OC evaluation campaigns for which system outputs are publicly available.

The first data set comes from the Replab 2013 reputational polarity task \cite{replab2013}. It consists of 61 companies with 1,500 tweets each; tweets are annotated as positive, negative, or neutral for the company's reputation. 

All the other five datasets are sentiment analysis subtasks from SemEval for which system outputs are available online: SemEval-2015 task 10A (1680 samples, 13 systems), task 10B (8985 samples, 51 systems) and task 10C (3097 samples, 11 systems)  \cite{semeval15-T10}; and SemEval-2014, tasks 9A (2392 samples, 48 systems) and 9B (2396 samples, 7 systems).  
All these tasks contain three categories. Given that SemEval tasks do not distribute samples in test cases, we emulate 10 test cases by dividing randomly the data sets into 10 partitions  in order to compute UIR. 

Table~\ref{tab:exp1} (right part) shows the results. $\CEM^\mathtt{ORD}$ is the top performer in four datasets, and the second best (with a minimal difference of $0.01$ with respect to the best metric) in the other two. The non-logarithmic version of $\CEM^\mathtt{ORD}$ is, again, worse than the logarithmic version in all cases (except one, SemEval 2014 task 9A, where they both give the same result).  
 
 Some metrics are able to achieve a high coverage in some data sets, but not in a consistent manner. For instance, Kappa maximizes the coverage in the last dataset in the table, but achieves an extremely low result for RepLab. In general, the table also shows that the relative coverage performance of metrics varies depending on the dataset characteristics.

Finally, we also computed metrics robustness in terms of Spearman correlation between system rankings produced by the metric for topics (or data set partition) pairs in the  campaigns. The highest robustness (0.57) is achieved by $\CEM^\mathtt{ORD}$, Accuracy and F-measure; and the lowest robustness (0.49) is achieved by Accuracy with 1 and Macro Average MAE.  $\CEM^\mathtt{ORD}$ is more robust than its non-logarithmic version   $\CEM^\mathtt{ORD}_{flat}$ (0.57 vs 0.55), again supporting the use of the information-theoretic logarithmic formula.

\section{Conclusions}
\label{sec:con}

Our findings can be summarized as follows: (i)~metrics commonly used for Ordinal Classification problems are highly heterogeneous and, in general, inconsistent with the notion of ordinal scale in Measurement Theory; (ii)~the notion of closeness between classes can be modelled in terms of Measurement Theory and Information Theory and particularized for different scales; and (iii)~our proposed Ordinal Closeness Evaluation Measure ($\CEM^{\mathtt{ORD}}$) is the only one that satisfies all desirable formal properties, it is as robust as the best state-of-the-art metrics, and it is the one that better captures the different quality aspects of \OC problems in our experimentation, with both  synthetic and naturalistic datasets. 

From a methodological perspective, the evidence that we have presented covers the four approaches pointed out  in~\citet{Amigo-18}: we have compared metrics in terms of desirable formal properties to be satisfied (theoretic top-down), we have generalized existing approaches (theoretic bottom-up), and we have compared effectiveness on human assessed and on synthetic data (empirical bottom-up and top-down). Future work includes the application of $\CEM$ at scales other than the ordinal. 
\

Code to compute $\CEM$ will be available at
{\footnotesize \texttt{github.com/EvALLTEAM/EvALLToolkit}}.

\section*{Acknowledgements}

This research has been partially supported by grants {\em Vemodalen} (TIN2015-71785-R) and {\em MISMIS} (PGC2018-096212-B-C32) from the Spanish government, as well as by the Google Research Award \emph{Axiometrics: Foundations of Evaluation Metrics in IR}. 


\bibliography{OrdinalClassification}
\bibliographystyle{acl_natbib}

\cleardoublepage
 \newpage
  \begin{appendices}


\begin{figure*}[t]
    \centering
 
\noindent
\renewcommand\arraystretch{1.3}
\setlength\tabcolsep{0pt}
\begin{tabular}{c @{\hspace{0.4em}}l  @{\hspace{0.4em}}r @{\hspace{0.6em}}r @{\hspace{0.5em}}r  @{\hspace{0.5em}}r}
    \multirow{7}{*}{\rotatebox{90}{\parbox{1.6cm}{\bfseries system A}}} & 
    & \multicolumn{3}{l}{\bfseries ground truth} & \\
  & & \bfseries neg &  \bfseries neu & \bfseries pos & \bfseries total \\ \hline
    & \bfseries neg$_A$ & 5 & 5 & 7 & 17 \\
  & \bfseries neu$_A$ & 1 & 50 & 8 & 59 \\
   &\bfseries pos$_A$ & 4 & 5 & 15 & 24 \\ \hline
  & total & 10 & 60 & 30 & 100
\end{tabular}
\ \ \ 
\renewcommand\arraystretch{1.3}
\setlength\tabcolsep{0pt}
\begin{tabular}{c @{\hspace{0.4em}}l  @{\hspace{0.4em}}r @{\hspace{0.5em}}r @{\hspace{0.5em}}r  @{\hspace{0.5em}}r}
    \multirow{7}{*}{\rotatebox{90}{\parbox{1.6cm}{\bfseries system B}}} & 
    & \multicolumn{3}{l}{\bfseries ground truth} & \\
  & & \bfseries neg &  \bfseries neu & \bfseries pos & \bfseries total \\ \hline
  & \bfseries neg$_B$ & 7 & 12 & 4 & 23 \\
  & \bfseries neu$_B$ & 1 & 45 & 8 & 54 \\
   &\bfseries pos$_B$ & 2 & 3 & 18 & 23 \\ \hline
  & total & 10 & 60 & 30 & 100
\end{tabular}
\ \ \
\renewcommand\arraystretch{1.3}
\setlength\tabcolsep{0pt}
\begin{tabular}{c @{\hspace{0.4em}}l  @{\hspace{0.4em}}S @{\hspace{0.5em}}S @{\hspace{0.5em}}S}
 &  \multicolumn{4}{c}{\bfseries class proximity}  \\
  & & \multicolumn{1}{c}{\bfseries neg} &  \multicolumn{1}{c}{\bfseries neu} & \multicolumn{1}{c}{\bfseries pos}  \\ \hline
  & \bfseries neg & 4.32 & 0.62 & 0.07  \\
  & \bfseries neu &  1.32    & 1.74  & 0.74  \\
   &\bfseries pos &   0.23   & 0.42 & 2.74  \\ \hline
   \\
\end{tabular}

\newcommand{\IQ}[1]{-\log\frac{#1}{100}}

{\small
\[\arraycolsep=1.5pt\def\arraystretch{2.2}
\begin{array}{lll}
  \prox(\mbox{neg},\mbox{neg})=\IQ{10/2}=4.32   &\prox(\mbox{neg},\mbox{neu})=\IQ{10/2+60}=0.62  & \prox(\mbox{neg},\mbox{pos})=\IQ{10/2+90}=0.07\\
  \prox(\mbox{neu},\mbox{neg})=\IQ{60/2+10}= 1.32&  \prox(\mbox{neu},\mbox{neu})=\IQ{60/2}= 1.74& \prox(\mbox{neu},\mbox{pos})=\IQ{60/2+30}= 0.74\\ 
  \prox(\mbox{pos},\mbox{neg})=\IQ{30/2+60+10}=0.23& \prox(\mbox{pos},\mbox{neu})=\IQ{30/2+60}=0.42 &
  \prox(\mbox{pos},\mbox{pos})=\IQ{30/2}= 2.74\\
\end{array}
\]

\newcommand{\total}{10*4.32+60*1.74+30*2.74}
\newcommand{\numA}{5*4.32+5*0.62+7*0.07+1*1.32+50*1.74+8*0.74+4*0.23+5*0.42+15*2.74}
\newcommand{\numB}{7*4.32+12*0.62+4*0.07+1*1.32+45*1.74+8*0.74+2*0.23+3*0.42+18*2.74}

\begin{align*}
 \CEM^\mathtt{ORD}(A,g)=& \frac{\numA}{\total}=0.71\\
 \CEM^\mathtt{ORD}(B,g)=& \frac{\numB}{\total}=0.76
\end{align*} 
}


   
    \caption{Example computation of $\CEM^\mathtt{ORD}$ values for two hypothetical systems A and B with respect to the same dataset. The first two tables represent the confusion matrices for both systems. The third table shows $prox(c_i,c_j)$ for the ground truth, according to the distribution of items in the negative, positive and neutral classes (10, 60 and 30, respectively). The rest of the equations illustrate how proximity values between classes are computed, and the resulting $\CEM^\mathtt{ORD}$ values for both systems.}
    \label{fig:example-calculation}
\end{figure*}


\section*{Appendix A. Example computation of $\CEM$}

Figure~\ref{fig:example-calculation} illustrates the computation of $\CEM$ for two systems (A and B) on the same ground truth with the three usual classes in sentiment analysis: negative, neutral, positive. The ground truth distribution is 10, 60 and 30 items, respectively, which is all the information needed to compute proximity between classes. Note that proximity of one class with respect to other is $-\log$ of the amount of items that lie between them (including all items in the ground truth class and half of the items in the system-predicted class) divided by the total number of items. The lowest score corresponds to the proximity between the two extreme cases (in the example, the negative and positive classes), because all items except half of the items in the system-predicted class lie between them, and therefore the $-\log$ value is minimal.

System A and System B in the figure both have the same accuracy ($0.70$), but system B receives a higher $\CEM^\mathtt{ORD}$ score (0.76 vs 0.71). The main reason is that system A makes more mistakes between distant classes (positive and negative). Another reason is that system A makes more positive/neutral than negative/neutral mistakes; and positive/neutral errors are more penalized by the metric than negative/neutral. The reason is that, together, the positive and neutral classes represent 90\% of the items in the dataset, and therefore are considered less close from an information-theoretic point of view.

%

\section*{Appendix B. Metric Properties Counter Examples}

Here we provide examples of how certain metrics fail to satisfy some of 
the properties proposed in the paper. 

\noindent
{\bf Ordinal Monotonicity}. Let us consider the set of categories $\Cat=\{1,2,3,4,5\}$. All classification metrics and correlation coefficients fail to satisfy {\em ordinal monotonicity}, given that for all of them: 
\begin{align*}
    \M((1,2,3),(3,4,5))=\M((2,3,4),(3,4,5)).
\end{align*}

But, according to the ordinal monotonicity property, the system output $(2,3,4)$ should receive a higher value than $(1,2,3)$, because all predicted classes are closer to the ground truth labels. 

\noindent
{\bf Ordinal Invariance} Pearson correlation, and every error minimization metric fails to satisfy {\em ordinal invariance}, given that for all of them:  
\begin{align*}
&\M((1,2,3),(3,4,5))\neq \\ &\M((f(1),f(2),f(3)),(f(3),f(4),f(5))
\end{align*}
being $f$, for instance, the strict (not linear) increasing function $f(x)=10+x^3$. 

\noindent
{\bf Imbalance}. According to the imbalance property,
\begin{align*}
\M((1,{\bf 2},2,3),&(1,1,2,3)) > \\ &\M((1,1,2,{\bf 2}),(1,1,2,3)).
\end{align*}

Metrics that do not satisfy this restriction are Accuracy $\left(\frac{3}{4},\frac{3}{4}\right)$, Accuracy with 1 $\left(1,1\right)$, MAE and MSE$\left(-\frac{1}{4},-\frac{1}{4}\right)$, cosine similarity $\left(0.973,0.979\right)$ and Pearson (0.85,0.9).

\section*{Appendix C. Proofs}
Here we provide proofs for the properties satisfied by metrics in our study. For the sake of brevity, we do not include formal complete proofs, but their explanations.

\noindent
{\bf Proof for closeness conditions at  ordinal scale}: Focusing on the ordinal scale, if $x$ is located between $y$ and $r$ ($y\le x\le r$ or $r\le x\le y$), then $|f(x)-f(r)|\le|f(y)-f(r)|$ for any strict increasing function $f$. In other case, that is, if $x<y \wedge x<r$ or $y<x \wedge r<x$ we can define a strict increasing function that invalidates $|f(x)-f(r)|\le|f(y)-f(r)|$. The reasoning for the strict case is similar. 

\noindent
{\bf Proof for $\CEM^\mathtt{ORD}$ properties:}
$\CEM^\mathtt{ORD}$ is computed over ordinal comparisons $(\cProxL{g(d')}{s(d)}{g(d)}{\mathtt{ORD}})$. By definition, closeness at ordinal scale is invariant under ordinal transformation. Therefore, 
$\CEM^\mathtt{ORD}$ is ordinal invariant. Monotonicity is also satisfied given that approaching the predicted category to the ground truth category necessarily reduces the amount of documents appearing in intermediate categories (provided there is no empty category in the gold standard), and therefore increases the similarity weight used by the metric. Finally, imbalance is also satisfied given that, being $g(d_i)=c_i$ and being $c_i$ and $c_j$ contiguous classes:
\begin{align*}
\CEM^\mathtt{ORD}&(g_{d_i\rightarrow c_j},g)-\CEM^\mathtt{ORD}(g,g)\\
&\propto
-\log\left(\frac{n_i+\frac{n_j}{2}}{N}\right)-\left(-\log\left(\frac{\frac{n_i}{2}}{N}\right)\right) .
\end{align*}
Therefore, 
\begin{align*}
&\M(g_{d_1\rightarrow c_2},g)-\M(g_{d_3\rightarrow c_2},g)\\
&\propto\M(g,g)-\log\left(\frac{n_1+\frac{n_2}{2}}{N}\right)-\left(-\log\left(\frac{\frac{n_1}{2}}{N}\right)\right)\\
&-\left(\M(g,g)-\log\left(\frac{n_3+\frac{n_2}{2}}{N}\right)-\left(-\log\left(\frac{\frac{n_3}{2}}{N}\right)\right)\right)\\
\propto& \log\left(\frac{\frac{n_1}{2}(n_3+\frac{n_2}{2})}{(n_1+\frac{n_2}{2})\frac{n_3}{2}}\right) ,
\end{align*}
which is larger than 0 whenever $n_1>n_3$.
\end{appendices}

\end{document}